\documentclass{article}

\usepackage{microtype}
\usepackage{graphicx}
\usepackage{amsthm}
\usepackage{mdframed}
\usepackage{booktabs} 

\usepackage{hyperref}
\usepackage{comment}

\usepackage[accepted]{icml2026preprint}

\usepackage{amsmath}
\usepackage{amssymb}
\usepackage{mathtools}
\usepackage{amsthm}

\usepackage{multirow}
\usepackage{subcaption}

\usepackage[capitalize]{cleveref}

\crefname{definition}{Definition}{Definitions}
\Crefname{definition}{Definition}{Definitions}

\theoremstyle{plain}
\newtheorem{theorem}{Theorem}[section]

\theoremstyle{definition}
\newtheorem{definition}[theorem]{Definition}

\theoremstyle{remark}

\newtheorem{example}[theorem]{Example}

\surroundwithmdframed[
    linewidth=1pt,
    linecolor=black,
    innertopmargin=3mm
]{example}

\crefname{definition}{Definition}{Definitions}
\Crefname{definition}{Definition}{Definitions}
\crefname{theorem}{Definition}{Definitions}
\Crefname{theorem}{Definition}{Definitions}

\usepackage[textsize=tiny]{todonotes}
\usepackage{researchpack}

\icmltitlerunning{Relational Linear Properties in Language Models: An Empirical Investigation}

\usepackage[dvipsnames]{xcolor}

\newcommand{\changed}[1]{{\color{black} #1}}

\definecolor{darkred}{RGB}{147, 17, 97}
\definecolor{midred}{RGB}{238, 132, 103}
\definecolor{midblue}{RGB}{117, 152, 246}
\definecolor{darkblue}{RGB}{55, 64, 124}

\newcommand{\cat}{\mathbin{\smallfrown}}
\newcommand{\LRE}{\texttt{LRE}\xspace}
\newcommand{\SVM}{\texttt{SVM}\xspace}
\newcommand{\method}{\texttt{KL-RP}\xspace}
\newcommand{\acronym}{KL-based Relational Probe}

\newcommand{\truth}{\textsc{Truth}\xspace}
\newcommand{\lang}{\textsc{Lang}\xspace}
\newcommand{\subj}{\textsc{Subj}\xspace}
\newcommand{\tense}{\textsc{Tense}\xspace}

\newcommand{\llama}{\texttt{Llama-3.1}\xspace}
\newcommand{\gemma}{\texttt{Gemma-2}\xspace}
\crefformat{equation}{Equation~#2#1#3}
\Crefformat{equation}{Equation~#2#1#3}

\crefmultiformat{equation}
  {Equations~#2#1#3}
  { and~#2#1#3}
  {, #2#1#3}
  {, and~#2#1#3}

\Crefmultiformat{equation}
  {Equations~#2#1#3}
  { and~#2#1#3}
  {, #2#1#3}
  {, and~#2#1#3}

\begin{document}

\twocolumn[
\icmltitle{Relational Linear Properties in Language Models:\\ An Empirical Investigation}



\icmlsetsymbol{equal}{*}

\begin{icmlauthorlist}
\icmlauthor{Giovanni Valer}{yyy,unibo,unipi}
\icmlauthor{Luigi Gresele}{comp}
\icmlauthor{Marco Bronzini}{equal,yyy,sch}
\icmlauthor{Emanuele Marconato}{equal,sch}
\end{icmlauthorlist}

\icmlaffiliation{yyy}{Fondazione Bruno Kessler, Trento, Italy}
\icmlaffiliation{comp}{University of Copenhagen, Copenhagen, Denmark}
\icmlaffiliation{sch}{University of Trento, Trento, Italy}
\icmlaffiliation{unibo}{University of Bologna, Bologna, Italy}
\icmlaffiliation{unipi}{University of Pisa, Pisa, Italy}

\icmlcorrespondingauthor{Giovanni Valer}{gvaler@fbk.eu}

\icmlkeywords{Language Models, Probing, Linear Properties, Latent Representation, Relational Linearity}

\vskip 0.3in
]



\printAffiliationsAndNotice{\icmlEqualContribution} 


\begin{abstract}
    Linear properties are ubiquitous in the representations of language models; 
    however, testing them 
    experimentally
    remains a challenging task. 
    \changed{This work focuses on \emph{relational linearity}: 
    the hypothesis that, for a fixed relation (\eg \textit{``plays''}), the unembedding of an object (\eg \textit{``trumpet''}) can be predicted from the embedding of its subject (\eg \textit{``Miles Davis''}) by a linear map.} 
    \changed{We present an experimental method to test the formulation of
    relational linearity by}~\citeauthor{marconato2024all}~\citeyearpar{marconato2024all}.  
    \changed{Specifically, we introduce a probing method, based on Kullback-Leibler divergence, to evaluate this property and examine its variation across layers and paraphrased relational queries.}
    It is also more efficient than previous work; for example, it avoids the crude Jacobian approximations used in Linear Relational Embeddings by \citet{hernandez2024linearity}.
    Our findings across four datasets show that relational linearity varies across models, exhibits layer-wise patterns consistent with prior observations about linguistic information in model representations, and is differently affected by changes in how the relation is phrased.
    
\end{abstract}

\section{Introduction}

How do large language models encode knowledge about the world, specifically about its entities and their relations, in their latent representations?
%
%
A growing line of work approaches this question by studying whether semantic properties of entities and relations are reflected in linear structure in the representation spaces of neural language models~\citep{mikolov2013distributed, elhage2022toy, Marks2023TheGO, park2023linear, engels2024not, wu2025axbench}.
Such linear structure has been argued to support both the interpretation of a model's internal computations~\citep{geiger2021causal} and the control of model behavior through activation steering~\citep{Marks2023TheGO, turner2023steering, wu2024reft}.\footnote{
\cref{sec:related_work} expands the discussion of these linear properties and reviews prior work on linear and relational properties of language models, as well as layer-wise interpretability.
}


\begin{figure*}[t]
    \centering

    \begin{tabular}{lc|cr}
        &
        \texttt{\llama} &
        \texttt{\gemma} \\
        \rotatebox{90}{
        \hspace{3em}
        $\vf(\vs)$   
        \hspace{6.5em}
        $\vf(\vs \cat \vq)$
        }
        &
        \includegraphics[height=17.5em]{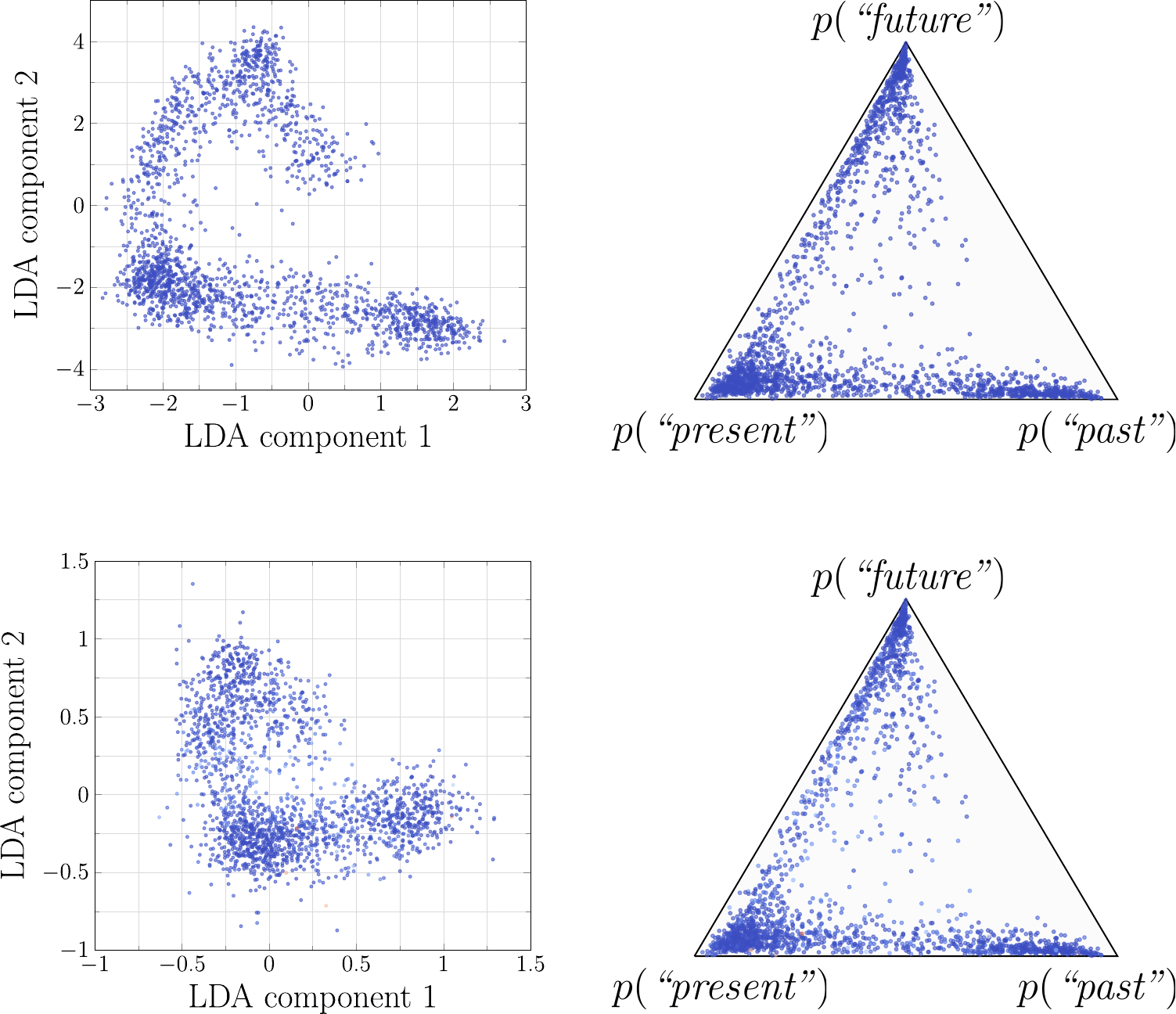}
        &
        \includegraphics[height=17.5em]{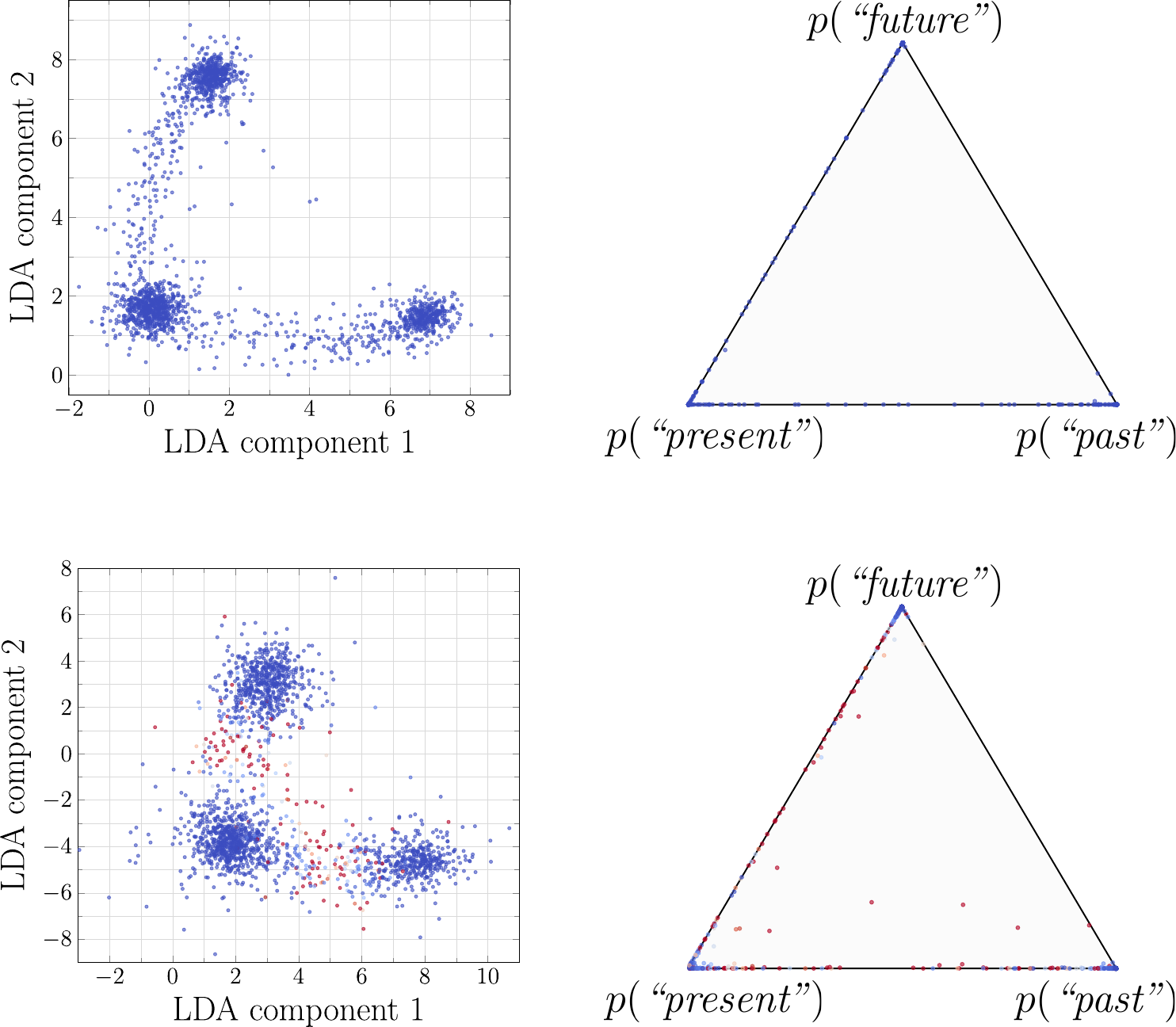}
        &  \raisebox{0.25\height}{\includegraphics[height=12em]{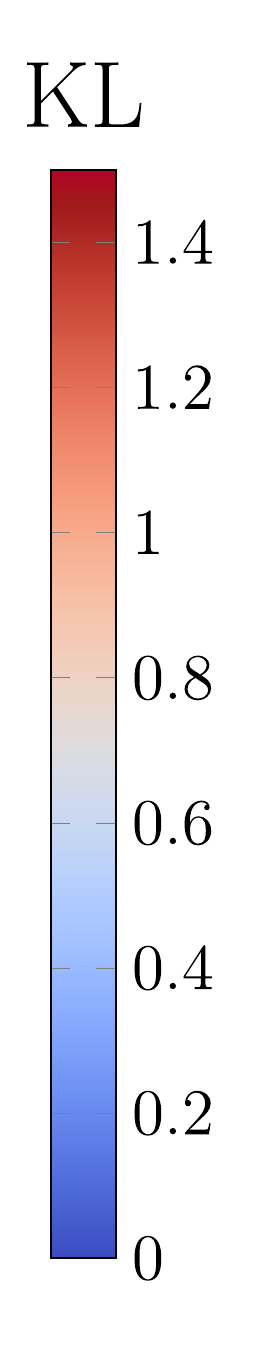}}  
    \end{tabular}
    
    \caption{
    \textbf{Relational linearity of final-layer representations for the \tense dataset}.
    For \llama{} (two left columns) and \gemma{} (two right columns), we compare the
    embeddings computed from contexts concatenated with the query ${\vq=\textit{``What is the tense of the previous sentence?''}}$ 
    \(\vf(\vs \cat \vq)\), with embeddings obtained by \acronym{} (\method) on the context-only embeddings \(\vf(\vs)\), for contexts \(\vs\) from the \tense dataset~\citep{ayman2024englishtense}. 
    The top row shows the reference embeddings \(\vf(\vs \cat \vq)\), while the bottom row shows the corresponding \method{} embeddings computed from \(\vf(\vs)\). 
    For each model, the left panels show two-dim LDA projections of the embeddings, and the right panels show the next-token probabilities for \textit{``present''}, \textit{``past''}, and \textit{``future''} on the probability simplex.
    Points are coloured by their KL divergence to the reference next-token distribution induced by \(\vf(\vs \cat \vq)\), with reference KL set to zero and values above \(1.5\) clipped for visualization. 
    For \llama{}, \method{} largely preserves both the embedding geometry and the induced next-token distributions, yielding \(d_{\mathrm{KL}} \approx 0.02\) on the test set. 
    For \gemma{}, the relation to the reference embedding geometry is weaker, and the induced next-token distributions differ more 
    from the reference, yielding \(d_{\mathrm{KL}} \approx 0.23\).
    } 
    \label{fig:main-fig}
\end{figure*}

%
Relational linearity in language models~\citep{paccanaro2002learning, hernandez2024linearity, marconato2024all} 
investigates whether relations between text sequences are encoded linearly in model representations. 
In subject-relation-object triples, or in context-query-answer triples, this means asking whether the object/answer associated with a relation/query can be recovered by a linear map from the representation of the subject/context alone. 
For example, for the subject \textit{``Jimi Hendrix''} and the relation \textit{``played at''}, the question is whether the unembedding of \textit{``Woodstock''} (object) can be predicted linearly from the representation of \textit{``Jimi Hendrix''} (subject), without appending the relation to the subject.
While relational linearity was detected in small datasets with annotated triplets and by resorting to linear expansions of transformer computations \citep{hernandez2024linearity, chanin2024identifying}, an extensive experimental evaluation of this property is currently missing.

We contribute to fill this gap and further investigate 
relational linearity in large language models, analyzing the degree to which it arises in their latent representations using a novel probing method based on Kullback-Leibler (KL) divergence.
\emph{Our main contribution is to introduce
an experimental procedure that permits evaluating relational linear properties}, 
extending~\citeauthor{marconato2024all}~\citeyearpar{marconato2024all}.
%
Specifically, our probing procedure only requires accessing the model's latent representations both when it is prompted with (a) the context and the query of interest, and with (b) the context only.
With this,
we investigate the presence of relational linearity in language models through a \textit{novel KL-based linear probe} (\method), and compare it with an adaptation of the Linear Relational Embedding (\LRE) method proposed by \citet{hernandez2024linearity}.
Computationally, our approach avoids crude Jacobian approximations of previous works~\citep{hernandez2024linearity} and is thus more efficient.
In experiments, our methods uncovers relational linear properties in language model representations that are missed by previous approaches, as we show in~\cref{sec:q1}.
We perform experiments using two language models: a medium-sized model (Llama-3.1, \citealp{grattafiori2024llama}) with 8B parameters, and a smaller 2B-parameter model (Gemma 2, \citealp{team2024gemma}); and we
study \changed{their internal activations} 
with different sizes: $4096$ for Llama, and $2304$ for Gemma.
Our experimental analysis reveals the following: 
\begin{enumerate}[leftmargin=2em,itemsep=1pt,topsep=1pt, label=\alph*)]
    \item language models encode relational properties in a largely linear manner, especially for tense  (\cref{fig:main-fig}) and truthfulness, with model-dependent or weaker effects for language and subjectivity (Q1; \Cref{sec:q1});
    \item relational linearity varies across layers, with surface features emerging in early layers and more abstract properties peaking in middle layers (Q2; \Cref{sec:q2});
    \item changing the way the query is phrased produces varied effects on linear probing (Q3; \Cref{sec:q3}).
\end{enumerate}




\section{Preliminaries} \label{sec:preliminaries}

{We denote by \(\calY\) a finite vocabulary of tokens. 
Textual inputs are elements \(\vx \in \calX\), where \(\calX\) is a set of finite ordered sequences of tokens from \(\calY\). 
We denote the operation of concatenation of token sequences by ``$\cat$''; for example, if tokens \(y_1,y_2,y_3\) are rendered as \textit{``A''}, \textit{``perfect''}, and \textit{``circle''}, 
then \(y_1 \cat y_2 \cat y_3 
\) is rendered as \textit{``A perfect circle''}.}

{

Given a textual input \(\vx \in \calX\), autoregressive language models compute a sequence of hidden states through \(L>1\) transformer layers. 
We denote by
$\vf^{(\ell)}(\vx) \in \bbR^d$, with $\ell \in \{1,\ldots,L\}$,
the hidden state at layer \(\ell\) used to predict the next token,\footnote{
Each transformer layer produces hidden states at all input positions; \(\vf^{(\ell)}(\vx)\) denotes the final-position hidden state used for next-token prediction. We consider only these layer outputs, not internal components such as attention heads or MLP blocks.
} 
and write \(\vf := \vf^{(L)}\) for the final-layer \emph{embedding}. 
We use the term \emph{representation} broadly to refer to such hidden states, covering both intermediate-layer hidden states and final-layer embeddings. 
The model also has an output \emph{unembedding} map \(\vg:\calY \to \bbR^d\), which projects hidden states into the output vocabulary by associating each token \(y \in \calY\) with an unembedding vector.
Together, \(\vf\) and \(\vg\) define the next-token distribution:
\begin{equation}
    \label{eq:model-family}
    p_{\vf,\vg}(y \mid \vx) = \frac{
        \exp( \vf(\vx)^\top \vg(y)  )
    }{
        \sum_{y' \in \calY} \exp( \vf(\vx)^\top \vg(y')  )
    } \, .
\end{equation}
}

\subsection{Relational Linearity}

Following the setting by \citet{paccanaro2002learning}, we consider textual strings that can be decomposed as triplets $\vs \cat \vq \cat y$, where $\vs, \vq \in \calX$ and $y \in \calY$. 
These can be either \textit{subject}-\textit{relation}-\textit{object} triplets, \eg $\vs=$\textit{``David Gilmour''}, $\vq=$\textit{``plays the''}, and $y=$\textit{``guitar''}, or, more generally, as \textit{context}-\textit{query}-\textit{answer} triplets, 
\eg $\vs=$\textit{``Today I'm happy.''}, $\vq=$\textit{``Is it a positive statement?''}, and $y=$\textit{``Yes''}.\footnote{
The definition is 
not restricted to well-formed sentences: 
both the query $\vq$ and sequences $\vs$ could be arbitrary strings,
such as \textit{``@!as1smf kqw\#ù''}} 
\changed{We will refer to $\vs$ as to the \textit{context} (or the \textit{subject}), to $\vq$ as the \textit{query} (or the \textit{relation}), and to $y$ as the \textit{answer} (or the \textit{object}).}

Relational linearity \citep{hernandez2024linearity, marconato2024all} describes the property under which it is possible to linearly decode the (last-layer) representation 
of the context concatenated with the query, $\vf(\vs \cat \vq)$, from the representation of the context, $\vf(\vs)$, alone. 
We mainly follow the setup described in \citeauthor{marconato2024all}~\citeyearpar{marconato2024all} and analyze \textit{linear relational probing},\footnote{
\changed{\citet{marconato2024all} introduce a more general framework for relational linear properties of which our \cref{def:linear-relational-probing} is a special case. 
We provide the general definition in \cref{def:relational-linearity}. 
}
}
that is, when relational linearity holds for a subset $\calY_P \subseteq \calY$ of specific tokens related to query $\vq$.
\changed{
As with \citeauthor{hernandez2024linearity}~\citeyearpar{hernandez2024linearity}, we extend it to all layers of language models.
Formally, let the conditional distribution restricted to tokens in $\mathcal{Y}_P$ defined by
\[
\textstyle
 p(y \mid \cdot \, ; \calY_P) := p(y \mid \cdot) \left/ \sum_{y' \in \calY_P } p(y' \mid \cdot) \right..
\]
}
\vspace{-3mm}
\begin{definition}[Linear Relational Probing
]
    \label{def:linear-relational-probing}
    For a  query $\vq \in \calX$ and a subset of tokens $\calY_P \subseteq \calY$ of dimension $k = |\calY_P|$, a model $(\vf, \vg) $ can be \emph{linearly probed for $\vq$ on the subset $\calY_P$ at layer $\ell$} if there exist $k$ weights $\vw^y_\vq \in \bbR^{d}$ and biases $b_\vq^{y} \in \bbR$ such that, for all $y \in \calY_P$ and $\forall \vs \in \calX$,
    \[  
        p 
        (y \mid \vs \cat \vq ; \calY_P) 
        =
        \frac{
            \exp( \vw_\vq^{y, \top} 
            \vf^{(\ell)}
            (\vs) + b_\vq^y )
        }{
            \sum_{y' \in \calY_P} \exp( \vw_\vq^{y', \top} \vf^{(\ell)}(\vs) + b_\vq^{y'} )
        }
        \, .
    \label{eq:stronger-relational-linear-probing}
    \]
\end{definition}
The following example illustrates the setting:
\vspace{-1mm}
\begin{example} \label{example:linear-probing}
    Consider a query $\vq =$\textit{``What is the tense of the previous sentence?''} and consider three possible answers $\calY_P = \{  
    \textit{``past''}, 
    \textit{``present''}, 
    \textit{``future''} 
    \}$. 
    Take as context $\vs=$\textit{``Yesterday, we watched a movie''}, to which the language model assigns high-probability to \textit{``past''} token (with probability $1 -2\epsilon$) and vanishing probability to \textit{``present''} and \textit{``future''} (both with probability $\epsilon$) 
    \changed{when prompted with input $\vs \cat \vq$}.
    %
    
    The language model can be linearly probed for $\vq$ on $\calY_P$ \changed{at layer $\ell$} 
    if this probability distribution can be \textit{exactly} reconstructed 
    from the embeddings $\vf(\vs)$ with a linear classifier and softmax activation (according to \cref{eq:stronger-relational-linear-probing} and for multiple different contexts $\vs \in \calX$). 
\end{example}

Intuitively, \cref{def:linear-relational-probing} ensures that, for all prompted contexts, the probability distribution on the relevant next-tokens for a query can be obtained through a linear probe of the representations of the context alone.

\subsection{Relaxation of Linear Relational Probing}

Similarly to \citeauthor{hernandez2024linearity}~\citeyearpar{hernandez2024linearity}, we also evaluate a weaker version of linear relational probing that only considers the likeliest tokens. 
In this setting, 
we do not need to consider the full equivalence between distributions as in \cref{eq:stronger-relational-linear-probing}, but only the match among the likeliest tokens. This specializes \cref{eq:stronger-relational-linear-probing} to the following (see \cref{def:linear-relational-probing-weak} for a complete description):  
\[
    \argmax_{y \in \calY_P} p(y \mid \vs \cat \vq; \calY_P) 
    = \argmax_{y \in \calY_P} \big( \vw_\vq^{y, \top} \vf^{(\ell)}(\vs) + b_\vq^y \big) \, .
    \label{eq:weaker-version-relational-linear-probing}
\]
Notice that, for linear relational probing,  \cref{eq:stronger-relational-linear-probing} is a stronger condition than \cref{eq:weaker-version-relational-linear-probing}, as the former equation implies the latter, but  
the opposite direction does not hold. 
A model can satisfy weak linear relational probing and not linear relational probing when, for example, both 
the linear probe and the query-prompted reference distribution $p(y \mid \vs \cat \vq)$ agree on the most-likely token, but with different probabilities. 
E.g., \textit{``past''} in Example \ref{example:linear-probing} may be the likeliest token for both the linear relational probe and the query-prompted reference, but with probability $(1 - \epsilon)$ for the former and $(1 - 2\epsilon)$ for the latter.  


\section{Estimating Relational Linearity} \label{sec:estimating_relational_linarity}

In this section, we present our main experimental setting and discuss different methods to estimate relational linearity. 
We first remark one key aspect of relational linearity that allows for straightforward evaluation (\cref{sec:considerations-relational-linearity}), then we introduce the main setup for testing (\cref{sec:experimental-setup}), and finally we present the different probes that we test (\cref{sec:methods-extended}). 


\begingroup
\hypersetup{hidelinks}
\subsection{Practical Implication of \cref{def:linear-relational-probing}}
\label{sec:considerations-relational-linearity}
\endgroup


We notice that
the notion of relational linear probing and its variant we discussed so far
only depend on how the language model defines the next-token distribution.
Upon selecting a query $\vq$ to be tested and the set of tokens $\calY_P$, one only needs to access the model representations $\vf(\vs)$ and $\vf(\vs \cat \vq)$, for multiple contexts $\vs \in \calX$, and the embossed probabilities on $\calY_P$ to validate if \cref{def:linear-relational-probing} and/or a weaker version hold. 
This has two immediate consequences.

On the downside, the likeliest tokens for the model might be nonsensical w.r.t. the prompted pair of context and query.  
E.g., for the context $\vs=$\textit{``I watched a movie.''} and query $\vq=$\textit{``What is the tense of the sentence?''}, one would expect higher likelihood on the token $y_1=$\textit{``past''} rather than $y_2=$\textit{``present''}. 
However, the language model might not be able to attribute high-likelihood on $y_1$ and predict $y_2$ \changed{or another token at random; this might happen for poorly trained language models.}
Whether this happens or not, it is irrelevant for \cref{def:linear-relational-probing} (and variants) to hold. 
{We observe this in one scenario when \llama is prompted to classify subjective \textit{vs} objective statements (see \cref{fig:layer_analysis}, third column). 
}
\changed{For example, while \llama struggles to correctly recognize subjective statements, middle layers show that the model's (wrong) answers can be approximately retrieved in a linear subspace.  
%
}

Moreover, model probabilities might result in trivial scenarios, where probabilities are collapsed in a small region of the simplex. E.g., if probabilities $p(y \mid \vs \cat \vq) = \mathrm{const}$ for all choices in contexts $\vs \in \calX$ and tokens in  $\calY$,\footnote{
A ``tautology" in \citeauthor{marconato2024all}~\citeyearpar[Definiton 23]{marconato2024all}. 
} \cref{def:linear-relational-probing} is easily satisfied by setting the weights $\vw^y_\vq$ to zero and properly tuning the biases $b^y_\vq$. 
We propose to detect these trivial scenarios introducing a \textbf{collapse on the simplex score} ($\mathrm{CSS}$) that leverages 
the normalized Shannon entropy 
$\mathrm{H}_k(\cdot)$%
\footnote{It uses a base-$k$ logarithm, where $k$ is the token count in $\mathcal{Y}_P$.} on a dataset of model probabilities $\{p(\cdot \mid \vs_i \cat \vq)\}_{i=1}^N$, that is %
\[
    \textstyle
    \mathrm{CSS} := 1 - 
    {\mathrm{H}_k \big( \overline{p(\cdot \mid \vs_i \cat \vq)} } \big) + \overline{ \mathrm{H}_k(p(\cdot \mid \vs_i \cat \vq))} \, ,
    \label{eq:triviality-score}
\] 
where $\overline{\cdot}$ denotes the average over the dataset. $\mathrm{CSS}$ ranges in $[0,1]$, where 
high values indicate a collapse towards $p(y \mid \vs \cat \vq) = \mathrm{const}$, whereas low values indicate that model probabilities span multiple regions of the simplex. \\

On the upside, estimating relational linearity does not require accessing to predefined and well-constructed triplets $(\vs, \vq, y)$ used in previous works \citep[\eg][]{hernandez2024linearity, chanin2024identifying}. 
This benefits a more thorough investigation upon deciding: (i) the dataset of contexts, (ii) the specific queries, and (iii) their relevant tokens. 
%
We discuss next how to make use of this effectively.

\subsection{Setup}
\label{sec:experimental-setup}


We assume to have access to a set of contexts $\calS = \{\vs_i \in \calX\}_{i=1}^N$. 
For each given query $\vq \in \calX$, 
we design a prompt $\tilde \vq$ to elicit the model for being more predictive of the relevant tokens $\calY_P$ chosen for the query.
In this setting, we test a slight variation of  \cref{def:linear-relational-probing} where the order of $\vs$ and $\tilde \vq$ is inverted and use an object-inducing string $\tilde \vt$:\footnote{
\eg a system prompt prepended to an LLM user’s input.}

\begin{example}
\label{example:prompt-setup}
    For the query $\vq=$\textit{``Does the previous sentence reflect a truthful statement?''} and tokens $\calY_P = \{\textit{``Yes''}, \textit{``No''} \}$, \changed{the system prompt is:}
\begin{center}
    $\tilde \vq =$\textit{``You are a linguistic classifier. Respond with only one word. Task: Identify the truthfulness of a sentence. Options: Yes, No.''}. 
\end{center}
Accordingly, the object-inducing string is given by
$\tilde \vt = \textit{``Truthfulness:''}$. 

\end{example}

From this, we evaluate 
the probability distribution of the model 
$p(y_j \mid \tilde \vq \cat \vs \cat \tilde \vt; \calY_P)$ for all relevant tokens $y_j \in \calY_P$.
This gives a dataset of input contexts embeddings and probabilities pairs
$\calD_\vq = \{\big( \vf(\vs_i), p(\cdot \mid \tilde \vq \cat \vs_i \cat \tilde \vt; \calY_P)  \big)  \}_{i=1}^N$. 

\textbf{Paraphrases}.
We note that some relations admit different query's surface forms (i.e., paraphrases), offering alternative ways to construct the prompt ($\tilde \vq$) and determine its relevant token set ($\calY_P$). 
The following illustrates an example of query paraphrasing for a binary relation:

\begin{example}
    As in example \ref{example:prompt-setup}, we consider the query $\vq=$\textit{``Does the previous sentence reflect a truthful statement?''} but now with tokens $\calY_P' = \{\textit{``True''}, \textit{``False''} \}$, 
    The system prompt we use in this setting is a paraphrase of $\tilde \vq$:
\begin{center}
    $\tilde \vq'  =$\textit{``You are a linguistic classifier. Respond with only one word. \textbf{Task: Classify a sentence as true or false}. Options: True, False.''}. 
\end{center}
The object-inducing string is
$\tilde \vt' = \textit{``Truthfulness:''}$.
\label{example:paraphrasing}
\end{example}

A natural question is whether different textual alternatives of a query exhibit comparable relational linearity, which we empirically evaluate in \cref{sec:empirical_verification} for two distinct relations. 

\subsection{Methods}
\label{sec:methods-extended}

To address the question of whether a language model encodes any relational linearity, we consider the following estimation methods.
First, we repurpose the technique of Linear Relational Embedding (\LRE) by \citeauthor{hernandez2024linearity}~\citeyearpar{hernandez2024linearity}.
In short,
\LRE estimates the weights and biases for linear relational probing 
by resorting to a first-order Taylor expansion of the context concatenated with the query. 
In practice, by choosing a small set of contexts (usually way smaller than $N$), the joint embedding $\vf(\tilde \vq \cat \vs \cat \tilde \vt)$ is approximated as:
\[
    \vf(\tilde \vq \cat \vs \cat \tilde \vt) \approx \beta \hat \vW_\vq \vf(\vs) + \hat \vb_\vq
\]
where $\beta \in \bbR$ is a fine-tuned term,  $\hat \vW_\vq$ and $\hat \vb_\vq$ are obtained by averaging the terms from first-order Taylor expansion on the subset of contexts.
\changed{
This is similar in spirit to estimating the broader notion of relational linearity (see \cref{def:relational-linearity}), but requires estimating the Jacobian, which can make the procedure computationally intensive and introduce additional numerical or estimation-related variability.}
For more details, we refer to \citeauthor{hernandez2024linearity}'s work \citeyearpar{hernandez2024linearity}.

%
{
\changed{Second, we introduce the \acronym{} (\method), specifically tailored to test relational linear probing. \method builds on a linear model, with $k$ weights $\vw_y \in \bbR^{d}$ and biases $b_y \in \bbR$, and is trained to align to the probability of the language model by minimizing the Kullback-Leibler (KL) divergence. 
\method defines a conditional distribution over the set of tokens in $\calY_P$ given by
}
\[
    p^{(\ell)}_{\vW, \vb} (y \mid \vs) = \frac{
        \exp( \vw_y^\top \vf^{(\ell)}(\vs) + b_y )
    }{
        \sum_{y' \in \calY_P} \exp( \vw_{y'}^\top \vf^{(\ell)}(\vs) + b_{y'} )
    } \, .
\]
}
The weights and biases are chosen by minimizing the KL divergence to the model conditional distribution, that is
\[
    d_\mathrm{KL} =  \frac{1}{N}
    \sum_{i=1}^N
    \sum_{y' \in \calY_P} p(y' \mid \tilde \vq \cat \vs_i \cat \tilde \vt) \log \frac{ p^{(\ell)}_{\vW, \vb} (y' \mid \vs_i) }{p(y' \mid \tilde \vq \cat \vs_i \cat \tilde \vt)} \, .
    \label{eq:kl-divergence-dataset}
\]


\section{Empirical Verification} \label{sec:empirical_verification}

We aim to answer the following research questions:
\begin{itemize}
    \item[\textbf{Q1}] Do language models encode any linear relational property, and to what degree?

    \item[\textbf{Q2}] How does relational linearity vary across layers of these models?

    \item[\textbf{Q3}] Are paraphrases of the queries similarly encoded in model representations? 
\end{itemize}
\Cref{sec:experimental-setting} describes our experimental setup, and \cref{sec:experiment-language-models-prompts} evaluates the next-token performance of language models under the proposed query prompt. We then address the three research questions in \cref{sec:q1,sec:q2,sec:q3}.

\subsection{Setting}
\label{sec:experimental-setting}

\textbf{Language Models}. We adopt two different language models:
    \llama{}-8B,\footnote{\href{https://huggingface.co/meta-llama/Llama-3.1-8B-Instruct}{huggingface.co/meta-llama/Llama-3.1-8B-Instruct}} which has 32 hidden layers and an embedding size of 4096.
    \gemma{}-2B,\footnote{\href{https://huggingface.co/google/gemma-2-2b-it}{huggingface.co/google/gemma-2-2b-it}} which has 26 hidden layers and embedding size of 2304.
%
For both models, we adopt the instruction-tuned version and apply 8-bit quantization \cite{dettmers2022gpt3}.


\textbf{Methods}. 
We investigate the linear relational properties encoded in the neural embeddings of both language models using different methods:
\begin{enumerate}[leftmargin=1.5em]

    \item Linear Relational Embedding (\LRE) is implemented according to~\citeauthor{hernandez2024linearity}~\citeyearpar{hernandez2024linearity}. See \cref{app:lre_setup}.


    \item \method is a linear layer that minimizes\footnote{The model is trained using Adam with a learning rate of $10^{-4}$.} the KL divergence in \cref{eq:kl-divergence-dataset}. Softmax is used to evaluate the final probability of the tokens.
    
    \item \texttt{Random} baseline defines the null hypothesis for relational linearity. It uses the \method method, but is trained with randomly permuted embeddings (see \Cref{eq:random_baseline}). 


\end{enumerate}

A further probing method, based on Support Vector Machines (SVMs), is implemented to test the relaxed variant of linear relational probing (\cref{eq:weaker-version-relational-linear-probing}); details and discussion of this are left to \cref{app:svm}.

\textbf{Metrics}. To assess relational linearity, we measure:
\begin{itemize}
    \item \texttt{F1}(GT): the linear probe F1-score to ground-truth labels of the task;
    \item \texttt{F1}(LLM): F1-score for the model's most likely tokens;
    \item KL divergence between LLM and linear probe predictions (\cref{eq:kl-divergence-dataset}). 
\end{itemize}
As sanity check, we measure the collapse on the simplex score (\cref{eq:triviality-score}) on model probabilities when concatenating contexts and queries.
Note that F1(GT) provides a further sanity check, since a low F1(GT) combined with good performance in predicting the LLM output (either a high F1(LLM) or a low $d_{KL}$) might suggest some bias of the LLM in next-token prediction.
This is especially important when comparing \method to LRE, as the two methods require distinct prompt templates, causing the LLM to behave differently.

\textbf{Datasets}.
In our experiments, we consider four sentence-level datasets, each one annotated with a distinct linguistic or semantic attribute. We investigate relational linearity by examining language, tense, subjectivity, and truthfulness. For each of these, we adopt a different dataset, so that we have the ground truth label of the given relation:

\begin{itemize}[leftmargin=1.5em]
    \item \lang \cite{papariello2021language}. Each sentence is annotated with the origin language, for a total of 20 distinct idioms (each one ranging from 3.9\% to 5.2\% of the total).
    For our experiments, we consider only a subset of the total data, including 9542 samples.
    
    \item \tense   \cite{ayman2024englishtense}. Here, sentences are written in English and are labelled according to their main verbal tense (either \textit{past}, \textit{present}, or \textit{future}).
    Labels are in proportion 36.9\% \textit{present}, 33.8\% \textit{future}, 29.3\% \textit{past}. The dataset includes $10$k samples.
    
    \item \subj \cite{clef-2025}. Sentences are labeled based on whether they express an opinion or a fact. This multilingual dataset 
    covers Arabic (37\%), German (23\%),  English (14\%), Italian (14\%), and Bulgarian (12\%). The dataset counts $8090$ samples. 
    
    \item \truth \cite{Marks2023TheGO}. Sentences are binary labelled on factual accuracy, indicating whether they reflect true or false statements. 
    The two classes 
    are perfectly balanced.
     All sentences 
    are written in English. The dataset contains $9123$ samples.
\end{itemize}

Input sentences exceeding 250 characters (about 1.6\% of the total) are excluded to avoid exceedingly long contexts and to reduce computational costs. 
Moreover, for all datasets, we use 80\% of the entries as a training split, and the remaining 20\% for evaluation.
All datasets are well-balanced across classes.
\cref{app:data} reports the descriptive statistics of the four datasets (\cref{tab:datasets}).

\textbf{Queries.}
\changed{For each dataset, we separately construct a textual query designed to elicit model outputs that match the dataset’s labels. For example, for the \lang relation, we use a system prompt with the query ``What is the written language?'', allowing $20$ possible token-level answers corresponding to the languages in the dataset (see also \cref{tab:data_samples}). Throughout the paper, we refer to each query by its dataset name; for example, the above is denoted as the \lang query.}

\begin{table}[!bp]
    \centering
    \caption{Performance of language models in predicting the ground truth label when prompted with the query appended.  
    It also reports the $\mathrm{CSS}$, and a majority class \texttt{baseline}. Best across columns are marked in \textbf{bold}.
    }
    \scriptsize
    \resizebox{\linewidth}{!}{
    \begin{tabular}{lcccccccc}
        \toprule
         & \multicolumn{2}{c}{\textbf{\lang}} 
         & \multicolumn{2}{c}{\textbf{\tense}} 
         & \multicolumn{2}{c}{\textbf{\subj}} 
         & \multicolumn{2}{c}{\textbf{\truth}} \\
         & F1 & $\mathrm{CSS}$ & F1 & $\mathrm{CSS}$ & F1 & $\mathrm{CSS}$ & F1 & $\mathrm{CSS}$ \\
        \midrule
        \texttt{Baseline}
            & $0.05$ & $-$ 
            & $0.64$ & $-$
            & $\mathbf{0.64}$ & $-$
            & $0.50$ & $-$ \\
                          
        {\llama}
            & $\mathbf{0.97}$ & $0.46$
            & $0.89$ & $0.57$
            & $0.63$ & $\mathbf{0.46}$
            & $\mathbf{0.86}$ & $0.35$ \\
            
        {\gemma}
            & $0.59$ & $\mathbf{0.22}$
            & $\mathbf{0.90}$ & $\mathbf{0.05}$
            & $0.41$ & $0.91$
            & $0.70$ & $\mathbf{0.22}$ \\
        \bottomrule
    \end{tabular}}
    \label{tab:llm_performance}
\end{table}

\subsection{Performance of language models on next-token prediction}
\label{sec:experiment-language-models-prompts}

Initially, we explore the capabilities of language models to correctly predict the ground truth label on the four datasets. 
Results are presented in \cref{tab:llm_performance}.
The majority class baseline provides a reference for comparing LLM capabilities.
The collapse score (\Cref{eq:triviality-score}) is also reported as a preliminary measure of how indicative subsequent probing results are of relational linearity; values near 1 suggest less meaningful probing outcomes. \changed{
We highlight that \gemma on the \subj query shows high collapse ($\approx 0.91$), 
implying the relation is almost trivially encoded, as model output probabilities are nearly collapsed into a narrow region of the simplex.
}


We observe that \llama outperforms \gemma across all datasets, with 
\tense with marginal difference (0.89 vs. 0.90).
However, it is worth noting that for \subj, both models perform worse than the majority-class baseline.
We can argue that predicting input's subjectivity is challenging, and even more so in a multilingual setting without additional context, as in this case.
Whereas on the \lang relation, \llama outperforms \gemma by a large margin (0.97 vs 0.59). 
This is expected, as the \gemma is mostly trained on English sentences, whereas \llama encounters multiple languages during training, 
More broadly, this performance disparity is likely attributable to the difference in model size (8B vs 2B).

Interestingly, \gemma prompted with queries of all relations exhibits more peaked probability distributions over the tokens than \llama (see \cref{fig:max-probability-histogram}, \changed{reported in \cref{sec:app-additional-experiments}}). 
This can also be observed in the rightmost simplex plots in \cref{fig:main-fig}. 
Since our KL-based probe (\method) is trained directly on LLM predictions instead of ground truth, these preliminary findings on model's output distribution have to be taken into account as they set different baselines for the two language models.

\subsection{Q1: Do language models encode any linear relational property?}
\label{sec:q1}

\begin{table*}[!t]
    \caption{\textbf{
    Comparison between \method, \LRE, and a \texttt{Random} baseline}. Results are obtained with representations at middle layers ($\ell = 16$ for \llama; $\ell = 13$ for \gemma). Best across columns are marked in \textbf{bold}.
    }
    \centering
    \scriptsize
    \begin{tabular}{llcccrcccrcccrccc}
         \toprule
         & &
         \multicolumn{3}{c}{\lang}
         & &
         \multicolumn{3}{c}{\tense}
         & &
         \multicolumn{3}{c}{\subj}
         & &
         \multicolumn{3}{c}{\truth}
         \\
         \midrule
        \multirow{5}{*}{\rotatebox{90}{\texttt{Llama-3.1}}}
         & & 
         F1(GT) & F1(LLM) & $d_\mathrm{KL}$ & &
         F1(GT) & F1(LLM) & $d_\mathrm{KL}$ & &
         F1(GT) & F1(LLM) & $d_\mathrm{KL}$ & &
         F1(GT) & F1(LLM) & $d_\mathrm{KL}$          \\
         \cline{3-5} 
         \cline{7-9}
         \cline{11-13}
         \cline{15-17} 
         \\
             & \texttt{Random} &
             $0.04$ & $0.05$ & $2.04$ & & 
             $0.28$ & $0.36$ & $0.57$ & &
             $0.41$ & $0.47$ & $0.49$ & &
             $0.50$ & $0.49$ & $0.49$   
         \\
             & \LRE &
             $0.06$ & $0.36$ & $0.51$ & & 
             $0.15$ & $0.20$ & $1.46$ & & 
             $0.28$ & $0.50$ & $\mathbf{0.02}$ & & 
             $0.84$ & $0.90$ & $0.25$   
        \\
             & \method &
             $\mathbf{0.98}$ & $\mathbf{0.98}$ & $\mathbf{0.06}$ & & 
             $\mathbf{0.89}$ & $\mathbf{0.95}$ & $\mathbf{0.02}$ & & 
             $\mathbf{0.63}$ & $\mathbf{0.88}$ & $0.10$ & & 
             $\mathbf{0.86}$ & $\mathbf{0.94}$ & $\mathbf{0.04}$   
         \\
         \midrule \midrule
         %
         %
         \multirow{5}{*}{\rotatebox{90}{\texttt{Gemma-2}}}
         & & 
         F1(GT) & F1(LLM) & $d_\mathrm{KL}$ & &
         F1(GT) & F1(LLM) & $d_\mathrm{KL}$ & &
         F1(GT) & F1(LLM) & $d_\mathrm{KL}$ & &
         F1(GT) & F1(LLM) & $d_\mathrm{KL}$          \\
         \cline{3-5} 
         \cline{7-9}
         \cline{11-13}
         \cline{15-17} 
         \\
             & \texttt{Random} &
             $0.03$ & $0.13$ & $6.16$ & & 
             $0.29$ & $0.30$ & $1.70$ & &
             $\mathbf{0.39}$ & $\mathbf{0.50}$ & $\mathbf{0.16}$ & &
             $0.37$ & $0.47$ & $0.82$   
         \\
             & \LRE &
             $0.01$ & $0.48$ & $\mathbf{0.45}$ & & 
             $0.47$ & $0.48$ & $0.88$ & & 
             $0.27$ & $0.48$ & $0.38$ & & 
             $\mathbf{0.80}$ & $\mathbf{0.92}$ & $\mathbf{0.10}$   
        \\
             & \method &
             $\mathbf{0.51}$ & $\mathbf{0.62}$ & $2.19$ & & 
             $\mathbf{0.89}$ & $\mathbf{0.95}$ & $\mathbf{0.23}$ & & 
             $\mathbf{0.39}$ & $\mathbf{0.50}$ & $0.26$ & & 
             $0.72$ & $0.90$ & $0.42$   
         \\
     \bottomrule
    \end{tabular}
    \label{tab:extended-table-all}
\end{table*}


We compare \method to \LRE and \texttt{Random} baseline. Due to the poor scalability of \LRE (requiring approximately $30$ mins per relation to obtain the linear approximation for a single layer) compared to \method (approximately $40$ mins per relation to validate the probe across all layers), we focus on middle layers of \llama and \gemma.     

Table \ref{tab:extended-table-all} shows the performance of different linear probing approaches across the four datasets.
We observe that \method shows consistently higher F1(LLM) scores and lower KL divergence compared to the random baseline, across \lang, \tense, and \truth.
Such results indicate that $\vf(\vs)$ linearly encodes at least some information about the queries.
In particular, this is more evident for \tense and \truth in both LLMs, with F1(LLM) scores above 0.90 and KL divergence lower than 0.50.
For the \lang relation, \gemma performs considerably worse than \llama (0.62 vs. 0.98), though it still exceeds the random baseline. 
This suggests that its latent representations retain some degree of linearity for this relation, yet less than \llama’s.
%
We observe different outcomes on \subj: \method shows less distinctive linearity for \llama with F1(LLM) equal to 0.88 and no degree of linearity for \gemma (0.50).

Concerning the results of \LRE, there are two cases where good performance in predicting the LLM output is combined with a very low F1(GT) score: \llama on \subj, and \gemma on \lang.
Upon further inspection, we find that in both cases they exhibit a strong bias toward specific tokens: “False” in the first case, and “Arabic” and “Swahili” in the second.
This can be caused by the different prompt template required by \LRE (see \cref{app:prompt}).
We observe results close to \method only in the \truth relation for both models, with slightly worse performance for \llama and slightly better performance for \gemma.
In all other relations, it was not possible to linearly approximate the model's behavior using a \LRE. 
This could suggest that the model’s vector space separates the binary property of input's truthfulness more clearly, making it easier to approximate with linear methods.
\LRE not only reduces probing capability by \emph{capturing only strongly-encoded relational linearity due to its approximation}, but is also \emph{highly sensitive to hyperparameter choices} (\eg $\beta$, \cref{app:lre_setup}), adding an additional degree of freedom and leading to inconsistent probing outcomes.
Overall, \LRE misses some linear relational properties detected by \method, thus we focus on our method for the next research questions.

\subsection{Q2: How does relational linearity vary across layers?}
\label{sec:q2}

\begin{figure*}[htp]
    \centering
    \includegraphics[width=0.98\linewidth]{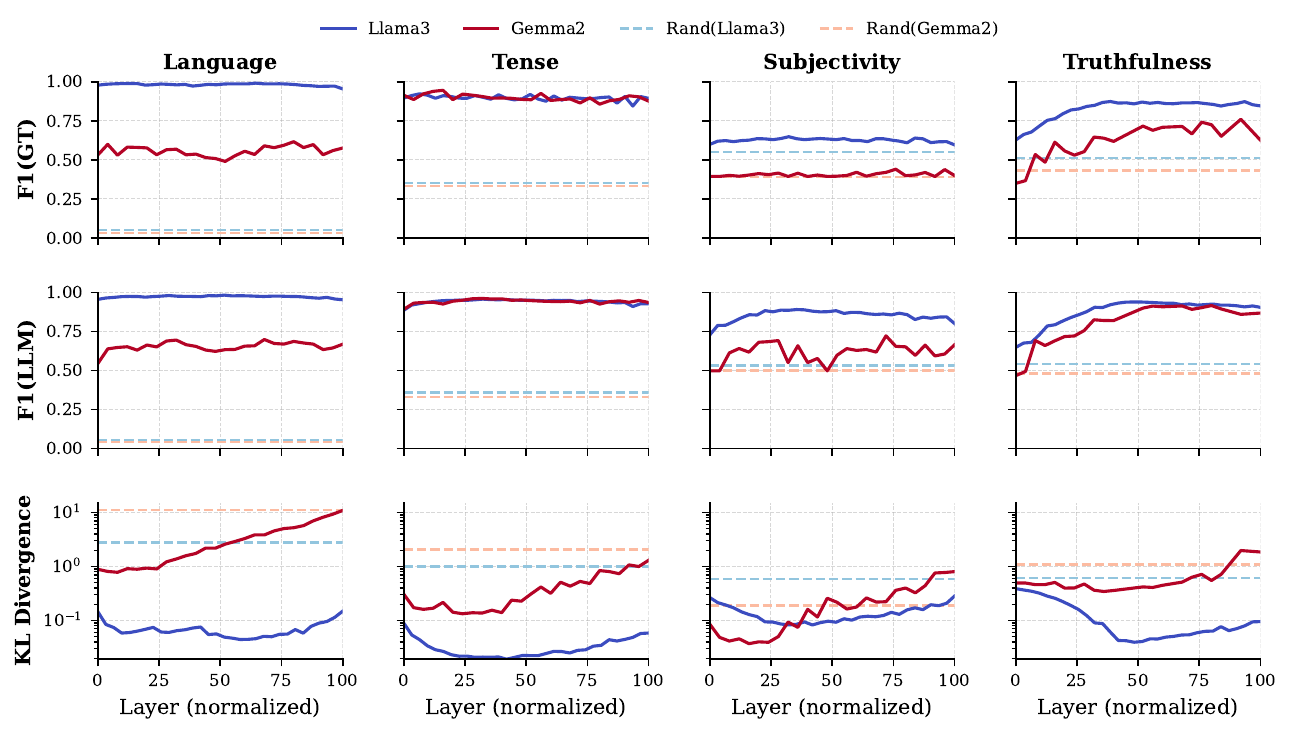}
    \caption{Relational linear probing across layers (normalized into the range 0--100), comparing \llama and \gemma.}
    \label{fig:layer_analysis}
\end{figure*}

Figure \ref{fig:layer_analysis} presents the layer-wise probing results with \method, including F1 scores and KL divergence. First, we analyze the F1 score.
When focusing on \llama, we observe distinct layer-wise dynamics between surface-level syntactic relations (\lang and \tense) and semantic and more abstract ones (\subj and \truth). 
Surface-level relations already achieve high F1 scores in early layers, whereas abstract relations show increasing F1(LLM), peaking in the middle layers. 
A similar pattern is also observed in \gemma, except for \subj, as expected due to 
its triviality score for this relation equal to $0.91$ (\cref{sec:experiment-language-models-prompts}; \cref{tab:llm_performance}).
These layer-wise findings align with experimental observations from prior work (\Cref{sec:related_work}; \eg \citet{tenney2019bert, cheng2025emergence}):
\emph{surface linguistic features emerge in early layers} (\ie \lang and \tense), while \emph{middle layers peak on abstract ones} (\ie \subj and \truth)

\begin{figure}[!bp]
    \centering
\includegraphics[width=0.95\linewidth]{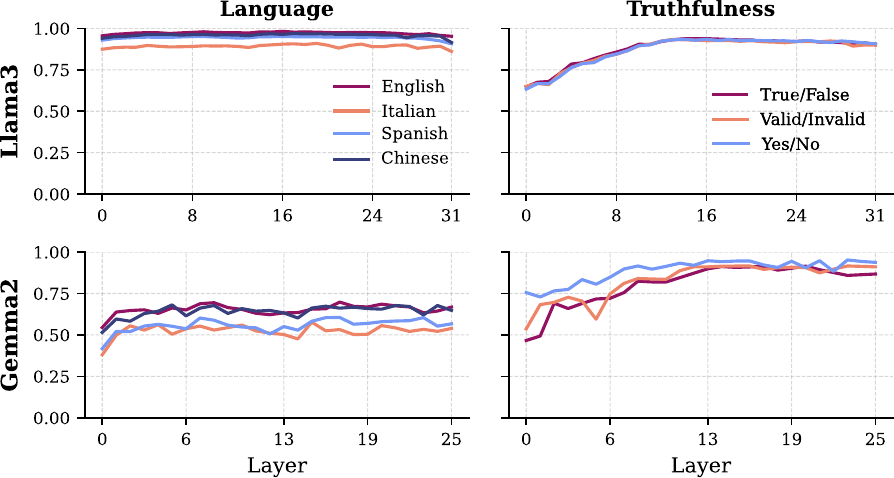} 
    \caption{Relational linear probing (\method) with different paraphrases.
    The graphs show \texttt{F1}(LLM) for both models. 
    }
    \label{fig:layer_analysis_paraphrases}
\end{figure}

\changed{About $d_\mathrm{KL}$, \llama representations show a sensible reduction to \texttt{Random} already at first layers, with noticeable improvements in middle layers (with major drops in the \tense and \truth datasets). 
Last-layers often increase in $d_\mathrm{KL}$ but not sensibly enough to downgrade F1 scores. Overall, middle layers for all datasets show high strong and weak relational linear probing for \llama.
Instead, \gemma representations in first and middle layers fare higher in $d_\mathrm{KL}$ but always below random chance. Here, KL scores are lower for \subj, however, for a trivial relation (with \texttt{Random} at $0.16$). Last-layers always degrade in $d_\mathrm{KL}$, showing that \gemma encode weak relational linearity at first and middle layers for linguistic datasets. 
}


\subsection{Q3: Are paraphrases of the queries similarly encoded in model representations?}
\label{sec:q3}

We evaluate how changes in the query’s surface form affect linear relational probing by varying the queries for: 
    (1) \lang: The query is translated 
    in four different languages: \textit{English}, \textit{Italian}, \textit{Spanish}, and \textit{Chinese}.
    (2) \truth: Three paraphrased formulations are evaluated, each corresponding to a distinct set of target tokens $\mathcal{Y}_P$: \textit{True}/\textit{False}, \textit{Valid}/\textit{Invalid}, and \textit{Yes}/\textit{No}.
    

\changed{We focus on these two relations 
to compare how paraphrases impact on {linguistic \textit{vs} abstract} queries.}
In particular, the binary \truth relation allows variation of the relevant target tokens, while the \lang relation enables the study of cross-lingual queries.
\changed{System prompts are in \cref{app:paraphrases}.}

\Cref{fig:layer_analysis_paraphrases} shows 
the layer-wise dynamics of \method in F1(LLM) tested with different paraphrases, with comprehensive results available in \cref{{app:paraphrases}}.
Changing the language of the query 
slightly impact on the probing results on both language models: for \gemma, Spanish and Italian lead to lower 
scores  ($-12.5$ on average)  with respect to English and Chinese across all layers consistently; while for \llama, only Italian  reduces probing performance of \method by 6.7.
For the paraphrases on \truth, 
\llama shows robustness to label variation, whereas \gemma exhibits substantial deviations in 
performance, with larger variations in earlier layers.
This suggests that: (i) the \emph{effect of surface-level query variations diminishes progressively} in \gemma, indicating that its middle layers may capture more abstract representations of this latent property,
and (ii) \llama appears to
\changed{encode relational linearity earlier},
as changes in query surface form have no effect across all layers.



\section{Limitations} \label{sec:limitations}
Our work presents some limitations. 
While our analysis provides the first and comprehensive empirical validation of the relational linearity formulation proposed by  \citet{marconato2024all}, we experiment with a slight variation where each query is structured using the model's system prompt (\Cref{sec:experimental-setup}). 
Despite this, the experimental results highlight that both language models we tested often encode strong relational linearity in low-dimensional subspaces across layers. 
We remark that this might not hold for the original formulation of relational linearity, \ie when considering embeddings $\vf(\vs \cat \vq)$ instead of $\vf(\tilde \vq \cat \vs \cat \tilde \vt)$, as the former might not induce well-defined answers as the latter. 
We intend to evaluate this formulation in future work.


While the proposed probe does not require annotated data, unlike prior work on relational linearity (\eg \citealp{hernandez2024linearity}), the annotations of our datasets (\eg written language in \lang) enable us to benchmark 
probe's performance against 
ground truth via F1 score.
However, this restricts the analysis of relational linearity to inputs that are likely to elicit well-defined answers to relational queries.
This narrows the evaluation of \cref{def:linear-relational-probing} to specific subsets of contexts, where relations are drawn from a bounded feature space.
A direction of improvement is to consider more generic datasets of contexts, \eg DBpedia~\citep{lehmann2015dbpedia}, enabling the assessment of relational linearity even when contexts do not yield well-defined answers to queries.
We plan to investigate this in future experiments.



\section{Conclusion} \label{sec:conclusions} 
This work investigates relational linearity in language models, providing an empirical validation of the formulation proposed by~\citet{marconato2024all}.
We introduce a novel probing approach based on Kullback-Leibler divergence to investigate it
for two distinct language models (Q1; \Cref{sec:q1}), evaluate its layer-wise dynamics (Q2; \Cref{sec:q2}), and the impact of paraphrases on relational queries (Q3; \Cref{sec:q3}). 
The experimental results indicate that (i) relational linearity holds in several cases, (ii) layer-wise patterns emerge consistent with prior evidence on linguistic dynamics (i.e., \lang appear in early layers, while \truth peaks in intermediate layers), and (iii) varying the query’s surface form has a slight effect on the results.


As future work,
we plan to study
relational linearity further, by choosing new queries,  investigate
emerging patterns in relational linearity caused by model distillation (\eg Llama 8B vs. 70B), and whether it can enable sensible steering.


\section*{Acknowledgments}

E.M. and M.B. acknowledge support from TANGO, Grant Agreement No. 101120763, funded by the European Union. Views and opinions expressed are however those of the author(s) only and do not necessarily reflect those of the European Union or the European Health and Digital Executive Agency (HaDEA). Neither the European Union nor the granting authority can be held responsible for them. M.B. has been also partially supported by the European Union’s Horizon Europe research and innovation program under grant agreement No. 101120237 (ELIAS).
L.G. was supported by the Pioneer Centre for AI, DNRF grant number P1.

\bibliography{references}

@inproceedings{roeder2021linear,
  title={On linear identifiability of learned representations},
  author={Roeder, Geoffrey and Metz, Luke and Kingma, Durk},
  booktitle={International Conference on Machine Learning (ICML)},
  pages={9030--9039},
  year={2021},
  organization={PMLR}
}

@article{park2023linear,
  title={The linear representation hypothesis and the geometry of large language models},
  author={Park, Kiho and Choe, Yo Joong and Veitch, Victor},
  journal={arXiv preprint arXiv:2311.03658},
  year={2023}
}

@article{marconato2024all,
  title={All or None: Identifiable Linear Properties of Next-token Predictors in Language Modeling},
  author={Marconato, Emanuele and Lachapelle, S{\'e}bastien and Weichwald, Sebastian and Gresele, Luigi},
    journal={International Conference on Artificial Intelligence and Statistics (AISTATS)},
  year={2025}
}

@inproceedings{park2024linear,
  title={The Linear Representation Hypothesis and the Geometry of Large Language Models},
  author={Park, Kiho and Choe, Yo Joong and Veitch, Victor},
  booktitle={International Conference on Machine Learning (ICML)},
  pages={39643--39666},
  year={2024},
  organization={PMLR}
}

@misc{papariello2021language,
  author = {Luca Papariello},
  title = {Language-Identification},
  howpublished = {\url{https://huggingface.co/datasets/papluca/language-identification}},
  type = {dataset},
  year = {2021},
  month = {November},
  note = {Accessed: 2025-27-06}
}

@article{Marks2023TheGO,
  title={The Geometry of Truth: Emergent Linear Structure in Large Language Model Representations of True/False Datasets},
  author={Samuel Marks and Max Tegmark},
  journal={ArXiv},
  year={2023},
  volume={abs/2310.06824},
  url={https://api.semanticscholar.org/CorpusID:263831277}
}

@InProceedings{clef-2025,
  author="Alam, Firoj
  and Stru{\ss}, Julia Maria
  and Chakraborty, Tanmoy
  and Dietze, Stefan
  and Hafid, Salim
  and Korre, Katerina
  and Muti, Arianna
  and Nakov, Preslav
  and Ruggeri, Federico
  and Schellhammer, Sebastian
  and Setty, Vinay
  and Sundriyal, Megha
  and Todorov, Konstantin
  and V., Venktesh",
editor="Hauff, Claudia
  and Macdonald, Craig
  and Jannach, Dietmar
  and Kazai, Gabriella
  and Nardini, Franco Maria
  and Pinelli, Fabio
  and Silvestri, Fabrizio
  and Tonellotto, Nicola",
title="The CLEF-2025 CheckThat! Lab: Subjectivity, Fact-Checking, Claim Normalization, and Retrieval",
booktitle="Advances in Information Retrieval",
year="2025",
publisher="Springer Nature Switzerland",
address="Cham",
pages="467--478",
isbn="978-3-031-88720-8",
}

@article{ayman2024englishtense,
  title={EnglishTense: A large scale English texts dataset categorized into three categories: Past},
  author={Ayman, Umme and Rahman, Md Hafizur and Islam, Md Shafiqul},
  journal={Present, Future tenses},
  year={2024}
}

@article{hernandez2024linearity,
      title={Linearity of Relation Decoding in Transformer Language Models}, 
      author={Evan Hernandez and Arnab Sen Sharma and Tal Haklay and Kevin Meng and Martin Wattenberg and Jacob Andreas and Yonatan Belinkov and David Bau},
      year={2024},
      eprint={2308.09124},
      archivePrefix={arXiv},
      primaryClass={cs.CL},
      url={https://arxiv.org/abs/2308.09124}, 
}

@article{paccanaro2002learning,
  title={Learning distributed representations of concepts using linear relational embedding},
  author={Paccanaro, Alberto and Hinton, Geoffrey E.},
  journal={IEEE Transactions on Knowledge and Data Engineering},
  volume={13},
  number={2},
  pages={232--244},
  year={2002},
  publisher={IEEE}
}

@article{elhage2021mathematical,
   title={A Mathematical Framework for Transformer Circuits},
   author={Elhage, Nelson and Nanda, Neel and Olsson, Catherine and Henighan, Tom and Joseph, Nicholas and Mann, Ben and Askell, Amanda and Bai, Yuntao and Chen, Anna and Conerly, Tom and DasSarma, Nova and Drain, Dawn and Ganguli, Deep and Hatfield-Dodds, Zac and Hernandez, Danny and Jones, Andy and Kernion, Jackson and Lovitt, Liane and Ndousse, Kamal and Amodei, Dario and Brown, Tom and Clark, Jack and Kaplan, Jared and McCandlish, Sam and Olah, Chris},
   year={2021},
   journal={Transformer Circuits Thread},
   note={https://transformer-circuits.pub/2021/framework/index.html}
}

@article{ferrando2024primer,
  title={A primer on the inner workings of transformer-based language models},
  author={Ferrando, Javier and Sarti, Gabriele and Bisazza, Arianna and Costa-Juss{\`a}, Marta R},
  journal={arXiv preprint arXiv:2405.00208},
  year={2024}
}

@article{engels2024not,
  title={Not all language model features are linear},
  author={Engels, Joshua and Liao, Isaac and Michaud, Eric J and Gurnee, Wes and Tegmark, Max},
  journal={arXiv e-prints},
  pages={arXiv--2405},
  year={2024}
}

@article{shu2025survey,
  title={A survey on sparse autoencoders: Interpreting the internal mechanisms of large language models},
  author={Shu, Dong and Wu, Xuansheng and Zhao, Haiyan and Rai, Daking and Yao, Ziyu and Liu, Ninghao and Du, Mengnan},
  journal={arXiv preprint arXiv:2503.05613},
  year={2025}
}

@inproceedings{wu2025axbench,
  title={AxBench: Steering LLMs? Even Simple Baselines Outperform Sparse Autoencoders},
  author={Wu, Zhengxuan and Arora, Aryaman and Geiger, Atticus and Wang, Zheng and Huang, Jing and Jurafsky, Dan and Manning, Christopher D and Potts, Christopher},
  booktitle={International Conference on Machine Learning},
  pages={67035--67080},
  year={2025},
  organization={PMLR}
}

@article{elhage2022toy,
  title={Toy models of superposition},
  author={Elhage, Nelson and Hume, Tristan and Olsson, Catherine and Schiefer, Nicholas and Henighan, Tom and Kravec, Shauna and Hatfield-Dodds, Zac and Lasenby, Robert and Drain, Dawn and Chen, Carol and others},
  journal={arXiv preprint arXiv:2209.10652},
  year={2022}
}

@article{wu2024reft,
  title={Reft: Representation finetuning for language models},
  author={Wu, Zhengxuan and Arora, Aryaman and Wang, Zheng and Geiger, Atticus and Jurafsky, Dan and Manning, Christopher D and Potts, Christopher},
  journal={Advances in Neural Information Processing Systems},
  volume={37},
  pages={63908--63962},
  year={2024}
}

@article{geiger2021causal,
  title={Causal abstractions of neural networks},
  author={Geiger, Atticus and Lu, Hanson and Icard, Thomas and Potts, Christopher},
  journal={Advances in neural information processing systems},
  volume={34},
  pages={9574--9586},
  year={2021}
}

@article{turner2023steering,
  title={Steering language models with activation engineering},
  author={Turner, Alexander Matt and Thiergart, Lisa and Leech, Gavin and Udell, David and Vazquez, Juan J and Mini, Ulisse and MacDiarmid, Monte},
  journal={arXiv preprint arXiv:2308.10248},
  year={2023}
}

@inproceedings{chanin2024identifying,
  title={Identifying linear relational concepts in large language models},
  author={Chanin, David and Hunter, Anthony and Camburu, Oana-Maria},
  booktitle={Proceedings of the 2024 Conference of the North American Chapter of the Association for Computational Linguistics: Human Language Technologies (Volume 1: Long Papers)},
  pages={1524--1535},
  year={2024}
}

@article{mikolov2013distributed,
  title={Distributed representations of words and phrases and their compositionality},
  author={Mikolov, Tomas and Sutskever, Ilya and Chen, Kai and Corrado, Greg S and Dean, Jeff},
  journal={Advances in neural information processing systems},
  volume={26},
  year={2013}
}

@inproceedings{cheng2025emergence,
title={Emergence of a High-Dimensional Abstraction Phase in Language Transformers},
author={Emily Cheng and Diego Doimo and Corentin Kervadec and Iuri Macocco and Lei Yu and Alessandro Laio and Marco Baroni},
booktitle={The Thirteenth International Conference on Learning Representations},
year={2025},
url={https://openreview.net/forum?id=0fD3iIBhlV}
}

@inproceedings{tenney2019bert,
  title={BERT rediscovers the classical NLP pipeline},
  author={Tenney, Ian and Das, Dipanjan and Pavlick, Ellie},
  booktitle={Proceedings of the 57th annual meeting of the association for computational linguistics},
  pages={4593--4601},
  year={2019}
}

@article{grattafiori2024llama,
  title={The {L}lama 3 herd of models},
  author={Grattafiori, Aaron and Dubey, Abhimanyu and Jauhri, Abhinav and Pandey, Abhinav and Kadian, Abhishek and Al-Dahle, Ahmad and Letman, Aiesha and Mathur, Akhil and Schelten, Alan and Vaughan, Alex and others},
  journal={arXiv preprint arXiv:2407.21783},
  year={2024}
}

@article{team2024gemma,
  title={Gemma 2: Improving open language models at a practical size},
  author={{Gemma Team} and Riviere, Morgane and Pathak, Shreya and Sessa, Pier Giuseppe and Hardin, Cassidy and Bhupatiraju, Surya and Hussenot, L{\'e}onard and Mesnard, Thomas and Shahriari, Bobak and Ram{\'e}, Alexandre and others},
  journal={arXiv preprint arXiv:2408.00118},
  year={2024}
}

@article{dettmers2022gpt3,
  title={Gpt3. int8 (): 8-bit matrix multiplication for transformers at scale},
  author={Dettmers, Tim and Lewis, Mike and Belkada, Younes and Zettlemoyer, Luke},
  journal={Advances in neural information processing systems},
  volume={35},
  pages={30318--30332},
  year={2022}
}

@article{sharkey2025open,
  title={Open problems in mechanistic interpretability},
  author={Sharkey, Lee and Chughtai, Bilal and Batson, Joshua and Lindsey, Jack and Wu, Jeff and Bushnaq, Lucius and Goldowsky-Dill, Nicholas and Heimersheim, Stefan and Ortega, Alejandro and Bloom, Joseph and others},
  journal={arXiv preprint arXiv:2501.16496},
  year={2025}
}

@article{rajendran2024causal,
  title={From causal to concept-based representation learning},
  author={Rajendran, Goutham and Buchholz, Simon and Aragam, Bryon and Sch{\"o}lkopf, Bernhard and Ravikumar, Pradeep},
  journal={Advances in Neural Information Processing Systems},
  volume={37},
  pages={101250--101296},
  year={2024}
}

@inproceedings{goyal2025causal,
  title={Causal Differentiating Concepts: Interpreting LM Behavior via Causal Representation Learning},
  author={Goyal, Navita and Daum{\'e} III, Hal and Drouin, Alexandre and Sridhar, Dhanya},
  booktitle={The Thirty-ninth Annual Conference on Neural Information Processing Systems},
  year={2025}
}

@article{lehmann2015dbpedia,
  title={Dbpedia--a large-scale, multilingual knowledge base extracted from wikipedia},
  author={Lehmann, Jens and Isele, Robert and Jakob, Max and Jentzsch, Anja and Kontokostas, Dimitris and Mendes, Pablo N and Hellmann, Sebastian and Morsey, Mohamed and Van Kleef, Patrick and Auer, S{\"o}ren and others},
  journal={Semantic web},
  volume={6},
  number={2},
  pages={167--195},
  year={2015},
  publisher={SAGE Publications Sage UK: London, England}
}

@inproceedings{abdou2021can,
  title={Can language models encode perceptual structure without grounding? a case study in color},
  author={Abdou, Mostafa and Kulmizev, Artur and Hershcovich, Daniel and Frank, Stella and Pavlick, Ellie and S{\o}gaard, Anders},
  booktitle={Proceedings of the 25th conference on computational natural language learning},
  pages={109--132},
  year={2021}
}

@article{hernandez2023inspecting,
  title={Inspecting and editing knowledge representations in language models},
  author={Hernandez, Evan and Li, Belinda Z and Andreas, Jacob},
  journal={arXiv preprint arXiv:2304.00740},
  year={2023}
}

@article{sakata2026linear,
  title={Linear Representations of Hierarchical Concepts in Language Models},
  author={Sakata, Masaki and Heinzerling, Benjamin and Ito, Takumi and Yokoi, Sho and Inui, Kentaro},
  journal={arXiv preprint arXiv:2604.07886},
  year={2026}
}

@inproceedings{wang2024locating,
  title={Locating and extracting relational concepts in large language models},
  author={Wang, Zijian and Whyte, Britney and Xu, Chang},
  booktitle={Findings of the Association for Computational Linguistics: ACL 2024},
  pages={4818--4832},
  year={2024}
}

@article{popovivc2026tracing,
  title={Tracing Relational Knowledge Recall in Large Language Models},
  author={Popovi{\v{c}}, Nicholas and F{\"a}rber, Michael},
  journal={arXiv preprint arXiv:2604.19934},
  year={2026}
}

@article{lv2024interpreting,
  title={Interpreting key mechanisms of factual recall in transformer-based language models},
  author={Lv, Ang and Chen, Yuhan and Zhang, Kaiyi and Wang, Yulong and Liu, Lifeng and Wen, Ji-Rong and Xie, Jian and Yan, Rui},
  journal={arXiv preprint arXiv:2403.19521},
  year={2024}
}

@inproceedings{merullo2024language,
  title={Language models implement simple word2vec-style vector arithmetic},
  author={Merullo, Jack and Eickhoff, Carsten and Pavlick, Ellie},
  booktitle={Proceedings of the 2024 Conference of the North American Chapter of the Association for Computational Linguistics: Human Language Technologies (Volume 1: Long Papers)},
  pages={5030--5047},
  year={2024}
}
\bibliographystyle{icml2026}

\newpage
\appendix
\onecolumn
\crefalias{section}{appendix}

\section{Additional Related Work} \label{sec:related_work}


\paragraph{Linear Properties of Language Models.} 
The latent representations of transformer-based models, also known as residual stream, is a high-dimensional vector space that aggregates the outputs of all hidden layers~\citep{elhage2021mathematical, roeder2021linear,park2024linear}.
Notably, these representations are hard to interpret due to their
high dimensionality and the superposition phenomenon \citep{sharkey2025open}.
More recently, a lot of interest lies in the possibility of identifying linearly encoded, human-interpretable features in these vectors spaces \citep{elhage2022toy, hernandez2023inspecting, Marks2023TheGO,engels2024not,park2024linear,rajendran2024causal,ferrando2024primer, goyal2025causal}. 
Linear features, or concepts, can be recovered using supervised learning by training probes on input's weak labels or resorting to Sparse AutoEncoders (SAEs)~\citep{shu2025survey}. 
Our work focuses on relational linearity, which can differ from linearly encoded concepts \citep{chanin2024identifying, wang2024locating, sakata2026linear}, and does not directly require any annotation step for probing or post-hoc alignment (as in SAEs).


\paragraph{Relational Properties.} 
\citet{paccanaro2002learning} introduce the notion of linear relational embeddings to model relations between entities as linear transformations in vector spaces, enabling the encoding of relations with distributed representations. 
\changed{The works by \citep{abdou2021can, hernandez2024linearity} provides a first empirical evidence that certain relational mappings in language models can be approximated by linear transformations. 
\citet{hernandez2024linearity} frame \LRE to inspect relational linear properties, 
testing it on curated datasets of annotated (subject, relation, object) triplets across 47 relations. Subsequent work by \citet{chanin2024identifying} invert \LRE{s} to derive concept direction vectors. 
This and subsequent works testing relational linear properties \citep{wang2024locating, merullo2024language, lv2024interpreting, popovivc2026tracing,sakata2026linear} likewise rely on annotated subject–object pairs or relation-type labels to supervise or validate the linear structure.
%
}

This work empirically investigates the formulation of relational linearity proposed by~\citeauthor{marconato2024all}~\citeyearpar{marconato2024all}, which is grounded in the mechanisms by which language models compute next-token probability distributions.
%
Our approach enables testing relational linearity by training probes to match the model's output probabilities, eliminating the need for object annotations entirely: only a dataset of contexts and a choice of query are required.

\paragraph{Layer-wise interpretability.} 

\citeauthor{tenney2019bert}~\citeyearpar{tenney2019bert} have detected hierarchical distributed features, where
lower layers capture syntax (e.g., POS tags), while higher layers encode semantics (e.g., relations). Similarly,
\citeauthor{cheng2025emergence}~\citeyearpar{cheng2025emergence}
have shown that language is represented on a low–intrinsic-dimension manifold, with a high-dimensional phase in early–middle model's hidden layers. 
Our layer-wise findings on relational linearity (\Cref{sec:q2}) reflect this observed progression from surface to abstract linguistic properties.

\section{Additional Experimental Details and Results}
\label{sec:app-additional-experiments}
\label[appendix]{app:data}

\begin{table}[ht]
    \centering
    \caption{Overview of the datasets after pre-processing.}
    \label{tab:datasets}
    \begin{tabular}{lcc}
        \toprule
        \textbf{Dataset} & \textbf{\# Samples} & \textbf{\# Classes} \\
        \midrule
        \lang & 9,542 & 20 \\
        \tense & 10,000 & 3 \\
        \subj & 8,090 & 2 \\
        \truth & 9,123 & 2 \\
        \bottomrule
    \end{tabular}
\end{table}


\begin{table*}[h]
    \centering
    \caption{Examples from the datasets with the target class, alongside the textual query used in our work. Note that we hereby present only the \textit{Task} within the whole $\tilde \vq$, which is followed by the relation-specific tokens $\calY_P$ (refer to \cref{example:prompt-setup} for a complete prompt example).}
    \label{tab:data_samples}
    \begin{tabular}{l p{5cm} p{7.5cm} l}
        \toprule
        \textbf{Relation} & \textbf{Sentence} & \textbf{Query task} & \textbf{Target} \\
        \midrule
        \lang & \textit{Een man zingt en speelt gitaar.} & Identify the language of a sentence. & Dutch \\
        \hline
        \tense & \textit{I will be going to the concert.} & Identify the grammatical tense of the sentence below. & Future \\
        \hline
        \subj & \textit{They’re probably right.} & Identify if a sentence is a fact or an opinion. & Opinion \\
        \hline
        \truth & \textit{The capital city of Laos is Vienna.} & Classify a sentence as true or false. & False \\
        \bottomrule
    \end{tabular}
\end{table*}

\subsection{Prompt design}
\label{app:prompt}
As introduced in \cref{sec:experimental-setup}, we adopt a structured prompt that consists not only in $(\vs \cat \vq)$, but rather in $(\tilde \vq \cat \vs \cat \tilde \vt)$. Since our experiments are conducted with two instruction-tuned models, we use the system prompt applying the chat template as follows:

\begin{example} \label{example:llama-prompt}
Prompt for \llama (\method).

\scriptsize \ttfamily \noindent
<|begin\_of\_text|><|start\_header\_id|>system<|end\_header\_id|>

You are a linguistic classifier. Respond with only one word.\\
Task: Classify a sentence as true or false.\\
Options: True, False.

<|eot\_id|><|start\_header\_id|>user<|end\_header\_id|>

Sentence: The capital city of Laos is Vienna.\\
Truthfulness:
<|eot\_id|><|start\_header\_id|>assistant<|end\_header\_id|>
\end{example}

\begin{example} \label{example:gemma-prompt}
Prompt for \gemma (\method).

\scriptsize \ttfamily \noindent
<bos><start\_of\_turn>user

Sentence: The capital city of Laos is Vienna.
Truthfulness:
<end\_of\_turn>

<start\_of\_turn>system

You are a linguistic classifier. Respond with only one word.\\
Task: Classify a sentence as true or false.\\
Options: True, False.
<end\_of\_turn>

<start\_of\_turn>model
\end{example}

Due to the different setup required by \LRE, the prompt template used for such method differs:
\begin{example} \label{example:llama-prompt-lre}
Prompt for \llama (\LRE).

\scriptsize \ttfamily \noindent
<|begin\_of\_text|><|start\_header\_id|>user<|end\_header\_id|>

You are a linguistic classifier. Respond with only one word.\\
Task: Classify a sentence as true or false.\\
Options: True, False.

Sentence: Oranges are not Naturally Orange.\\
Truthfulness: False

Sentence: Volkswagen Group operates in the industry of consumer durables.\\
Truthfulness: True

Sentence: Seventy-one is larger than ninety-seven.\\
Truthfulness: False

Sentence: The capital city of Laos is Vienna.\\
Truthfulness:
<|eot\_id|><|start\_header\_id|>assistant<|end\_header\_id|>
\end{example}

\begin{example} \label{example:gemma-prompt-lre}
Prompt for \gemma (\LRE).

\scriptsize \ttfamily \noindent
<bos><start\_of\_turn>user

You are a linguistic classifier. Respond with only one word.\\
Task: Classify a sentence as true or false.\\
Options: True, False.

Sentence: Oranges are not Naturally Orange.\\
Truthfulness: False

Sentence: Volkswagen Group operates in the industry of consumer durables.\\
Truthfulness: True

Sentence: Seventy-one is larger than ninety-seven.\\
Truthfulness: False

Sentence: The capital city of Laos is Vienna.\\
Truthfulness:
<start\_of\_turn>model
\end{example}

\subsection{LRE setup \& hyperparameters search}
\label[appendix]{app:lre_setup}
LREs have four hyper-parameters: $\ell$, the layer after which $\mathbf{s}$ is to be extracted; $n$, the number of examples used to estimate the approximation; $\beta$, the rescaling factor applied to the linear map $W_r$; and $\rho$, the rank of the inverse $W^\dagger$.
To ensure the scalability of the method, we set the hyper-parameters $\ell$, $\beta$, $\rho$ per LLM and relation via grid search on a subset of data (10\%), while $n$ is set to 5 as \citet{hernandez2024linearity} show that for most relations scores start plateauing after $n=5$.
Similarly, the values for the grid search are chosen according to their results, and are: $\ell \in \left\{ 6, 11, 16, 23, 32 \right\}$ for \llama, while $\ell \in \left\{ 5, 9, 13, 20, 26 \right\}$ for \gemma; $\beta \in \left\{ 0.5, 1.0, 1.5, 2.0, 2.5, 3.0, 3.5, 4.0, 4.5, 5.0 \right\}$; and $\rho \in \left\{ 8, 16, 32, 64, 100 \right\}$.
Table \cref{tab:lre_hyperparams} shows the best value found, in terms of F1(LLM), for each hyperparameter.
\begin{table}[ht]
    \caption{Best configuration for each model and relation.}
    \label{tab:lre_hyperparams}
    \centering
    \begin{tabular}{lccccc}
    \toprule
    & & \lang & \tense & \subj & \truth \\
    \midrule
    \multirow{3}{*}{\llama} & $\ell$ & $16$ & $16$ & $6$ & $16$ \\
     & $\beta$ & $5.0$ & $5.0$ & $5.0$ & $2.5$ \\
     & $\rho$ & $100$ & $100$ & $100$ & $64$ \\
    \midrule
    \multirow{3}{*}{\gemma} & $\ell$ & $13$ & $13$ & $13$ & $13$ \\
     & $\beta$ & $5.0$ & $2.5$ & $0.5$ & $3.5$ \\
     & $\rho$ & $64$ & $32$ & $32$ & $64$ \\
    \bottomrule
    \end{tabular}
\end{table}

\begin{figure}[ht]
    \centering
    
    \begin{subcaptiongroup}
        \begin{subfigure}{0.2\textwidth}
            \includegraphics[width=\linewidth]{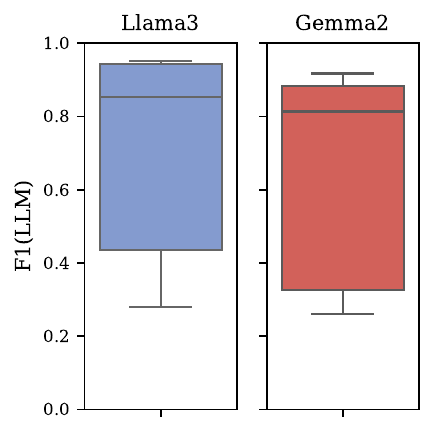}
            \caption{All layers}
        \end{subfigure}
        \begin{subfigure}{0.2\textwidth}
            \includegraphics[width=\linewidth]{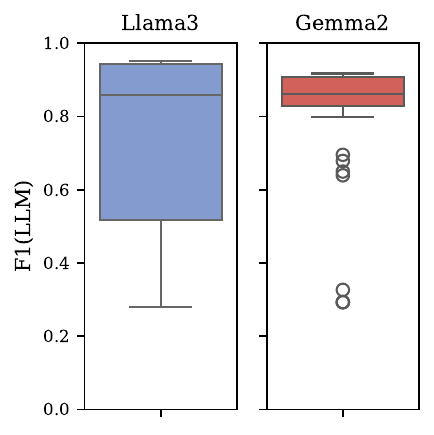}
            \caption{Middle layer}
        \end{subfigure}
    \end{subcaptiongroup}
    \caption{\textbf{Sensibility of \LRE when changing hyperparameters}.
    F1(LLM) scores on \truth using different configurations of \LRE hyper-parameters, \textbf{(a)} varying $\ell$, $\beta$, and $\rho$, or \textbf{(b)} with a fixed layer $\ell$.}
    \label{fig:lre-boxplot}
\end{figure}

Focusing on the relation leading to higher scores for \LRE, \ie \truth, \cref{fig:lre-boxplot} shows the results of the different hyper-parameter combinations.

\subsection{Randomly permuted embeddings for baseline}
The random baseline is defined by rearranging the embeddings with $\pi$, a random permutation of $\{1, \dots, N\}$, so that the randomized training data is:
\begin{equation}
        \calD_\vq^{\,\mathrm{rand}} = \left\{ \left( \vf(\vs_{\pi(i)}),\; p(\cdot \mid \tilde \vq \cat \vs_i \cat \tilde \vt; \calY_P) \right) \right\}_{i=1}^N
        \label{eq:random_baseline}
\end{equation}

\subsection{Probing with SVMs}
\label[appendix]{app:svm}
The Support Vector Machine (\SVM) with a linear kernel is trained to match the likeliest next-token of the language model for each pair $\tilde \vq \cat \vs \cat \tilde \vt$, according to \cref{eq:weaker-version-relational-linear-probing}. 
\changed{This rensembles the linear probing technique introduced by \citep{popovivc2026tracing}.}
Preliminary experiments involving this method have been conducted using \llama.
\cref{tab:svm} presents the results, showing how the two approaches reach close performance in terms of F1(LLM), but when considering $d_{KL}$, \method outperforms \SVM, as it is specifically trained to minimize the Kullback-Leibler (KL) divergence to the LLM predictions, rather than the most probable token only.

\begin{table*}[ht]
    \caption{Comparison of \method and \SVM.}
    \centering
    \scriptsize
    \begin{tabular}{lcccrcccrcccrccc}
         \toprule
         &
         \multicolumn{3}{c}{\lang}
         &
         &
         \multicolumn{3}{c}{\tense}
         &
         &
         \multicolumn{3}{c}{\subj}
         &
         &
         \multicolumn{3}{c}{\truth}
         \\
         \midrule
         &
         F1(GT) & F1(LLM) & $d_\mathrm{KL}$ & &
         F1(GT) & F1(LLM) & $d_\mathrm{KL}$ & &
         F1(GT) & F1(LLM) & $d_\mathrm{KL}$ & &
         F1(GT) & F1(LLM) & $d_\mathrm{KL}$          \\
         \cline{2-4} 
         \cline{6-8}
         \cline{10-12}
         \cline{14-16} 
         \\
         \texttt{Random} &
         $0.04$ & $0.05$ & $2.04$ & & 
         $0.28$ & $0.36$ & $0.57$ & &
         $0.41$ & $0.47$ & $0.49$ & &
         $0.50$ & $0.49$ & $0.49$  \\ 
     
         \texttt{SVM} &
         $\mathbf{0.99}$ & $\mathbf{0.98}$ & $0.79$ & &
         $0.88$ & $0.94$ & $0.76$ & &
         $\mathbf{0.65}$ & $0.85$ & $0.17$ & &
         $0.84$ & $0.93$ & $0.43$ \\
         
         \method &
         $0.98$ & $\mathbf{0.98}$ & $\mathbf{0.06}$ & &
         $\mathbf{0.89}$ & $\mathbf{0.95}$ & $\mathbf{0.02}$ & &
         $0.63$ & $\mathbf{0.88}$ & $\mathbf{0.10}$ & &
         $\mathbf{0.86}$ & $\mathbf{0.94}$ & $\mathbf{0.04}$ \\
         
     \bottomrule
    \end{tabular}
    \label{tab:svm}
\end{table*}

\begin{figure}[t]
    \centering
    \begin{tabular}{cccc}
        \includegraphics[width=0.225\linewidth]{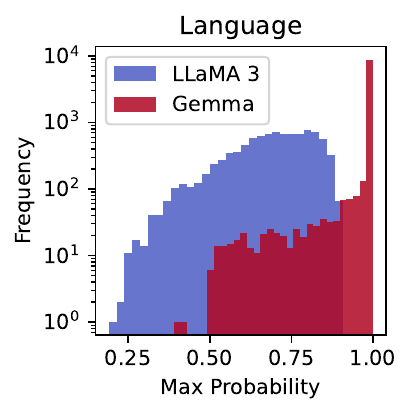}
         & 
        \includegraphics[width=0.225\linewidth]{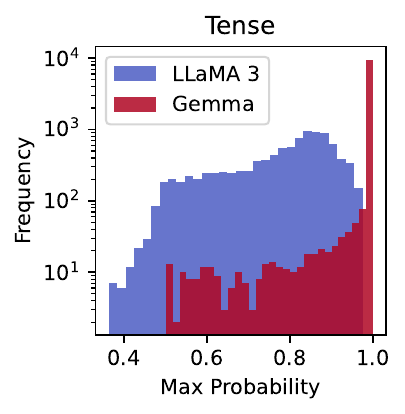} 
        \includegraphics[width=0.225\linewidth]{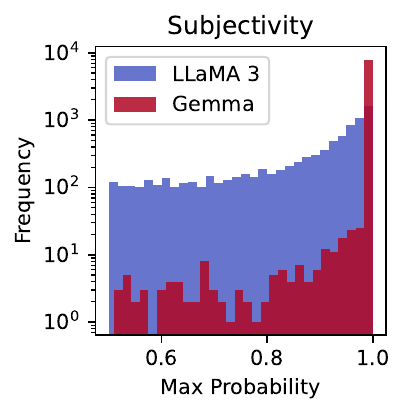}
         & 
        \includegraphics[width=0.225\linewidth]{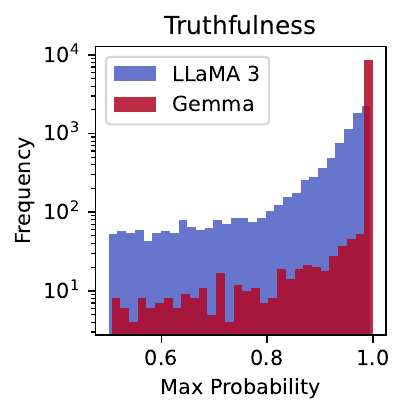} 
    \end{tabular}
    \caption{\textbf{Histograms of probabilities for \llama and \gemma}. The $x$-axis reports $\max p(y \mid \tilde \vq \cat \vs \cat \tilde \vt)$  when prompted with the query. 
    The $y$-axis is the frequency in log scale. We observe that \gemma tends to be more overconfident than \llama, collapsing output probabilities to high confidence regions two times more in order of magnitude than low confidence intervals.
    }
    \label{fig:max-probability-histogram}
\end{figure}

\section{Paraphrases}
\label[appendix]{app:paraphrases}

\cref{tab:paraphrases-results} shows the result of linear probing with \method across different paraphrases. We varied the language of the prompt for the \lang relation, experimenting with English $(i)$, Italian $(ii)$, Spanish $(iii)$, and Chinese $(iv)$. While for \truth we investigated different set of labels: \textit{True/False} $(i)$, \textit{Valid/Invalid} $(ii)$, and \textit{Yes/No} $(ii)$.
The comprehensive list of system prompts is provided in \cref{tab:paraphrases-prompts}.
\begin{table}[ht]
    \centering
    \caption{Result of different paraphrases. 
    All results refer to middle layer, \ie $\ell=16$ for \llama and $\ell = 13$ for \gemma.
    $-$ marks invalid measurements.
    }
    \scriptsize
    \begin{tabular}{c|ccccccc|ccccccc}
        \toprule
         \multicolumn{1}{c}{} & \multicolumn{7}{c|}{\llama} & \multicolumn{7}{c}{\gemma} \\
         \multicolumn{1}{c}{} & \multicolumn{3}{c}{\lang}
         &
         & \multicolumn{3}{c|}{\truth}
         & \multicolumn{3}{c}{\lang}
         &
         & \multicolumn{3}{c}{\truth}\\
         \midrule
         &
         F1(GT) & F1(LLM) & $d_\mathrm{KL}$ & &
         F1(GT) & F1(LLM) & $d_\mathrm{KL}$ &
         F1(GT) & F1(LLM) & $d_\mathrm{KL}$ & &
         F1(GT) & F1(LLM) & $d_\mathrm{KL}$          \\
         \cline{2-4} 
         \cline{6-8}
         \cline{9-11}
         \cline{13-15} 
         & \\

         $i$ & 
         $0.98$ & $0.98$ & $0.06$ & & 
         $0.86$ & $0.94$ & $0.04$ &   
         $0.51$ & $0.62$ & $2.19$ & & 
         $0.72$ & $0.90$ & $0.42$     
         \\
         $ii$ & 
         -- & $0.90$ & $0.08$ & & 
         -- & $0.93$ & $0.04$ &   
         -- & $0.51$ & $2.15$ & & 
         -- & $0.91$ & $0.39$     
         \\
         $iii$ & 
         -- & $0.95$ & $0.10$ & & 
         -- & $0.93$ & $0.04$ &   
         -- & $0.51$ & $2.28$ & & 
         -- & $0.92$ & $0.27$     
         \\
         $iv$ & 
         -- & $0.97$ & $0.07$ & & 
         -- & -- & -- &   
         -- & $0.65$ & $2.17$ & & 
         -- & -- & --     
         \\
         \bottomrule
    \end{tabular}
    \label{tab:paraphrases-results}
\end{table}

\begin{table}[!ht]
    \centering
    \scriptsize
    \caption{System prompt of each query paraphrase explored.}
    \begin{tabular}{c|p{7cm}|p{7cm}}
    \toprule
         & \lang & \truth \\
    \midrule
        $i$ & You are a linguistic classifier. Respond with only one word. Task: Identify the language of a sentence. Options: Arabic, Bulgarian, Greek, Hindi, Japanese, Russian, Thai, Urdu, Chinese, Dutch, Spanish, Italian, Turkish, Polish, Vietnamese, French, Portuguese, English, German, Swahili & 
        You are a linguistic classifier. Respond with only one word. Task: Classify a sentence as true or false. Options: True, False\\
    \midrule
        $ii$ & Sei un classificatore linguistico. Rispondi con una sola parola. Compito: Identifica la lingua di una frase. Opzioni: Arabo, Bulgaro, Greco, Hindi, Giapponese, Russo, Tailandese, Urdu, Cinese, Olandese, Spagnolo, Italiano, Turco, Polacco, Vietnamita, Francese, Portoghese, Inglese, Tedesco, Swahili &
        You are a linguistic classifier. Respond with only one word. Task: Classify the truthfulness of a sentence. Options: Valid, Invalid
\\
    \midrule
        $iii$ & Eres un clasificador lingüístico. Responde con una sola palabra. Tarea: Identifica el idioma de una oración. Opciones: Árabe, Búlgaro, Griego, Hindi, Japonés, Ruso, Tailandés, Urdu, Chino, Neerlandés, Español, Italiano, Turco, Polaco, Vietnamita, Francés, Portugués, Inglés, Alemán, Suajili  & 
        You are a linguistic classifier. Respond with only one word. Task: Classify the truthfulness of a sentence. Options: Yes, No
        \\
    \bottomrule
    \end{tabular}
    \label{tab:paraphrases-prompts}
\end{table}


\section{Other Definitions of Relational Linearity}
\changed{We collect here additional definitions of relational linearity by \citet{marconato2024all, hernandez2024linearity}. We first review \textit{relational linearity} of a query $\vq$, that is the capability of decoding the joint embedding $\vf(\vs\ cat \vq)$ from $\vf(\vs)$.
}
This may hold only on a linear subspace of the representation space, indicated as $\Gamma \subseteq \bbR^d$.
The formal definition follows:
\begin{definition}[Relational Linearity of $\vq$ in $\Gamma$ by \citealp{marconato2024all}]
\label{def:relational-linearity}
    For a query $\vq \in \calX$ and a subspace $\Gamma \subset \bbR^d$, a model $(\vf, \vg) $ \emph{linearly represents $\vq$ in $\Gamma$} if there exist a matrix $\vL_\vq \in \bbR^{d \times d} $ and a vector $\vd_\vq \in \bbR^d$ such that 
    \[  
        \forall \vs \in \calX , \;
        \vP_{\Gamma} \vf(\vs \cat \vq) 
        =\vP_{\Gamma} 
        \left( 
            \vL_\vq \vf(\vs) + \vd_\vq
        \right) \, .
    \]
\end{definition}
In other words, all the information of the representation $\vf(\vs \cat \vq)$ in the linear subspace $\Gamma$ is contained in (a subspace of) $\vf(\vs)$, modulo an affine transformation.  
Interestingly, this property can be declined in three settings that mirror empirical notions of linearly encoded concepts \citep{marconato2024all}, including \textit{linear relational subspaces}, \textit{linear relational probing} (\cref{def:linear-relational-probing}), and \textit{linear relational steering}. For a broader discussion, we refer the reader to \citep{marconato2024all}. 

Next, we provide the formal definition of \textit{weak linear relational probing}, accounting for argmax's of probability distribution of the model (measured with F1 scores). This is more aligned to the notion tested by \citet{hernandez2024linearity} in their experiments:

\begin{definition}[Weak Linear Relational Probing]
\label{def:linear-relational-probing-weak}
    For a  query $\vq \in \calX$ and a subset of tokens $\calY_P \subseteq \calY$ of dimension $k = |\calY_P|$, a model $(\vf, \vg) $ can be \emph{weakly linearly probed for $\vq$ on the subset $\calY_P$} if there exist $k$ weights $\vw^y_\vq \in \bbR^{d}$ and biases $b_\vq^{y} \in \bbR$ such that, for all $y \in \calY_P$ and $\forall \vs \in \calX$,
    \[
    \argmax_{y \in \calY_P} p(y \mid \vs \cat \vq; \calY_P) 
    = \argmax_{y \in \calY_P} \big( \vw_\vq^{y, \top} \vf(\vs) + b_\vq^y \big) \, .
    \]
\end{definition}

\end{document}